\documentclass{article} 
\usepackage{iclr2023_conference,times}

\usepackage{amsmath,amsfonts,bm}

\def\eqref#1{equation~\ref{#1}}

\def\1{\bm{1}}

\DeclareMathAlphabet{\mathsfit}{\encodingdefault}{\sfdefault}{m}{sl}
\SetMathAlphabet{\mathsfit}{bold}{\encodingdefault}{\sfdefault}{bx}{n}

\usepackage{hyperref}
\usepackage{url}
\usepackage{mathabx}

\usepackage{mathabx}
\usepackage{booktabs}
\usepackage{arydshln}
\usepackage{dsfont}
\usepackage{subcaption}
\usepackage{amsmath}
\usepackage{hyperref}
\usepackage{url}
\usepackage{graphicx,wrapfig,lipsum}

\usepackage{floatrow}
\newfloatcommand{capbtabbox}{table}[][\FBwidth]

\usepackage{blindtext}

\usepackage{multirow}
\usepackage[final]{changes}
\definechangesauthor[name={Anees Kazi}, color=blue]{AK}

\title{Projections of Model Spaces \\ for Latent Graph Inference}

\author{Haitz Sáez de Ocáriz Borde \thanks{Equal Contribution} \\
Oxford Robotics Institute,\\ University of Oxford\\
\texttt{haitz@oxfordrobotics.institute} \\
\And
Álvaro Arroyo  $^*$ \\
Oxford-Man Institute,\\ University of Oxford\\
\texttt{\{alvaro.arroyo\}@eng.ox.ac.uk} \\
\AND
Ingmar Posner \\
Oxford Robotics Institute,\\ University of Oxford\\
\texttt{\{ingmar.posner\}@eng.ox.ac.uk} \\
}

\begin{document}

\maketitle

\begin{abstract}

Graph Neural Networks leverage the connectivity structure of graphs as an inductive bias. Latent graph inference focuses on learning an adequate graph structure to diffuse information on and improve the downstream performance of the model. In this work we employ stereographic projections of the hyperbolic and spherical model spaces, as well as products of Riemannian manifolds, for the purpose of latent graph inference. Stereographically projected model spaces achieve comparable performance to their non-projected counterparts, while providing theoretical guarantees that avoid divergence of the spaces when the curvature tends to zero. We perform experiments on both homophilic and heterophilic graphs.
\end{abstract}
\vspace{-2mm}
\section{Introduction}
\vspace{-2mm}
Differential geometry has been widely used in physics; for instance, it has laid the mathematical foundations of the theory of general relativity as well as the gauge theory of quantum fields~\citep{Isham1989ModernDG}. Moreover, recent work within the machine learning community has started leveraging ideas which stem from differential geometry and topology to improve the performance of learning algorithms~\citep{Hensel2021ASO, Chamberlain2021, snns_barbero,barbero2022sheaf2}. Real-world data, which may manifest itself in high-dimensional observational space, can often be described via a lower-dimensional manifold according to the manifold hypothesis~\citep{Fefferman2013TestingTM}. Learning the underlying manifold which best describes the data can be a challenging task, but it is at the same time crucial for data modeling~\citep{Chen2022AutomatedDO,Du2020CompositionalVG,Du2021LearningSM}. We can hypothesize that even data which may only be observable in the form of disconnected point clouds, could be explained using a connected \textit{latent graph}, which is itself a discretization of a \textit{latent manifold}. Such latent graph can be applied to diffuse information through layers of Graph Neural Networks (GNNs) which are able to leverage geometric priors as inductive biases. At the same time, a better latent graph can be generated if an appropriate embedding space (or latent manifold) to describe the observed data is found. 

Several recent works have focused on optimizing the generation of latent graphs alongside the rest of the GNN model parameters~\citep{latentgraph,Kazi_2022,Cosmo2020LatentGraphLF} as well as on preprocessing the original input graph provided by the dataset \citep{Topping2021,diffusion_improves}. This idea is known as \textit{latent graph inference}. Other graph structure learning techniques include those by~\cite{Franceschi2019LearningDS,Chen2020IterativeDG,Jin2020GraphSL}. The discrete Differentiable Graph Module (dDGM) by~\cite{Kazi_2022} generates latent graphs based on Euclidean latent representations of the GNN layer activations. \cite{latentgraph} recently incorporated Riemannian geometry to this approach to generate more complex latent representations of the data, which need not be Euclidean. In particular, they used product manifolds of model spaces (the Euclidean plane, the hyperboloid, and the hypersphere) to approximate complex latent manifolds which help generate better latent graphs for improved downstream performance. In this work, we build upon this idea and introduce projections of model spaces to generate product manifolds. The use of these projections affords us improved theoretical properties, which guarantee that the model spaces will not diverge at zero curvature locations and that their distance and metric tensors will become Euclidean in these areas of the manifold. Furthemore, the use of these projected model spaces enables the creation of more diverse product manifold combinations while retaining computational tractability. 
\vspace{-2mm}
\section{Background}
\label{sec: Background}
\vspace{-2mm}
\subsection{Graph Neural Networks}
\vspace{-2mm}
GNNs are a class of artificial neural networks that have gained popularity due to their ability to exploit the geometric prior present in data as an inductive bias~\citep{bronstein2021geometric}. One key limitation of these models is that they require an input adjacency matrix that describes the connectivity of the graph on which they diffuse information. Furthermore, the initial connectivity structure is kept fixed even if it is unfriendly for message passing \citep{Topping2021}. GNNs generalize classical artificial neural networks to graphs $G=(\mathcal{V},\mathcal{E})$, where $\mathcal{V}$ represents the set of nodes and $\mathcal{E}$ the set of edges.  Each node in the graph has an associated $d$-dimensional feature vector $\mathbf{x}_v$. These can be grouped into a single matrix $\mathbf{X}$, and we can represent the set of all edges $\mathcal{E}$ via the adjacency matrix $\mathbf{A}$. GNN layers leverage these two matrices to transform features into a new set of latent features for each node $\mathbf{H}^{(l)} = f\left(\mathbf{H}^{(l-1)}, \mathbf{A}\right)$, most often sharing information in the 1-hop neighbourhood only; hence, stacking multiple layers allows the network to share information between more neighbors. Graph Convolutional Networks (GCNs)~\citep{Kipf2017SemiSupervisedCW} and Graph Attention Networks (GATs)~\citep{Velickovic2018GraphAN} are two examples of well-known GNN architectures.

\vspace{-2mm}
\subsection{Latent Graph Inference Leveraging Product Manifolds}
\label{sec: Latent Graph Inference Leveraging Product Manifolds}
\vspace{-2mm}

Latent graph inference, in the case of the dDGM module, consists on finding some measure of similarity between the latent node features to connect similar nodes and generate good latent graphs that are optimized based on a downstream task loss. In particular the probability of there being an edge between each node in the graph is $p_{ij}^{(l)}(\boldsymbol{\Theta}^{(l)})\ \propto  \exp(\varphi(f_{\boldsymbol{\Theta}}^{(l)}(\mathbf{x}_{i}^{(l)}),f_{\boldsymbol{\Theta}}^{(l)}(\mathbf{x}_{j}^{(l)});T)),$
where $\boldsymbol{\Theta}$ are the model parameters, $T$ is a learnable temperature hyperparameter, and most importantly, $\varphi$ is a measure of similarity between the features of each node in the graph. It is of key importance to find a meaningful and flexible measure of similarity $\varphi$ to improve downstream performance. This can be done using a distance metric based on the geodesic distance defined on a manifold $\mathcal{P}$. The probability of there existing an edge between node $i$ and $j$ in layer $l$ of the GNN is proportional to
\vspace{-1mm}
\begin{equation}
	p_{ij}^{(l)}\ \propto
 \exp(-T\mathfrak{d}_{\mathbb{P}}(f_{\mathbf{\Theta}}^{(l)}(\mathbf{x}_{i}^{(l)}),f_{\mathbf{\Theta}}^{(l)}(\mathbf{x}_{j}^{(l)}))),
\end{equation}
where $\mathfrak{d}_{\mathbb{P}}$ is the distance geodesic function between two points on the product manifold, and $\text{exp}(x)=e^{x}$ in this context. Based on this we can sample the edges of our latent graphs $\mathcal{E}^{(l)}(\mathbf{X}^{(l)};\mathbf{\Theta}^{(l)},T,k)=\{(i,j_{i,1}),(i,j_{i,2}),...,(i,j_{i,k}):i=1,...,N\},$ where $k$ is the number of sampled connections using the Gumbel Top-k trick. These edges are used to generate latent adjacency matrices $\mathbf{A}^{(l)}$, as described in~\citet{Kazi_2022}.

A manifold is a topological space in which all of its points have neighbourhoods that are homeomorphic to distinct open subsets of Euclidean hyperplanes. In particular, we define a Riemannian manifold $(\mathcal{M},g)$ as a real and differentiable manifold $\mathcal{M}$ for which each tangent space has an inner product $g$. This is called the Riemannian metric and varies smoothly between points on the manifold, meaning that it is continuous. The three so-called model spaces with constant curvature are the Euclidean plane, $\mathbb{E}^{d_{\mathbb{E}}}_{K_{\mathbb{E}}}=\mathbb{R}^{d_{\mathbb{E}}}$, where the curvature $K_{\mathbb{E}}=0$; the hyperboloid, $\mathbb{H}^{d_\mathbb{H}}_{K_{\mathbb{H}}}=\{\mathbf{x}_p \in \mathbb{R}^{d_\mathbb{H}+1} : \langle \mathbf{x}_p,\mathbf{x}_p \rangle_{\mathcal{L}} = 1/K_{\mathbb{H}} \},
$ where $K_{\mathbb{H}}<0$ and $\langle \cdot,\cdot \rangle_{\mathcal{L}}$ is the Lorentz inner product; and the hypersphere, $\mathbb{S}^{d_\mathbb{S}}_{K_{\mathbb{S}}}=\{\mathbf{x}_p \in \mathbb{R}^{d_\mathbb{S}+1} : \langle \mathbf{x}_p,\mathbf{x}_p \rangle_{2} = 1/K_{\mathbb{S}} \},$ where $K_{\mathbb{S}}>0$ and $\langle \cdot,\cdot \rangle_{2}$ is the standard Euclidean inner product. Each of these model spaces has an associated exponential maps to project the Euclidean outputs of GNN layers to each of the manifolds, and distance functions to compute distances between points on the manifolds, which are discussed in~\cite{latentgraph}. Product manifolds can be constructed using the Cartesian product $\mathcal{P} = \bigtimes_{i=1}^{n_{\mathcal{P}}} \mathcal{M}_{K_{i}}^{d_i},$ of other manifolds with curvature $K_{i}$ and dimensionality $d_i$. Note that in this work the dimensionality is set to 2 for all model spaces, but the curvature is learned during training to improve downstream performance. The distance between two points on the manifold can be computed by aggregating the distance contributions from each manifold composing the product manifold as follows
\begin{equation}
	\mathfrak{d}_{\mathcal{P}}(\mathbf{\overline{x}}_{p_1},\mathbf{\overline{x}}_{p_2})=\sqrt{\sum_{n=1}^{N}\mathfrak{d}_{i}\left(\mathbf{\overline{x}}_{p_1}^{(i)},\mathbf{\overline{x}}_{p_2}^{(i)}\right)^{2}},
\end{equation}
where the overline denotes that the points $\mathbf{\overline{x}}_{p_1}$ and $\mathbf{\overline{x}}_{p_2}$ have been adequately projected on to the product manifold using the exponential map before computing the distances, and $\mathfrak{d}_{i}$ is the distance function in each of the individual model spaces used to construct the overall product manifold. Note that the exponential map to project points from an Euclidean tangent plane, $\mathcal{T}_{p}\mathcal{P}$, onto the product manifold, $\mathcal{P}$, which is the concatenation of the exponential maps for each of its model space components. In turn, the \textit{manifold signature} is the number of model spaces used to generate product manifolds as well as their individual dimensionalities.
\vspace{-2mm}
\section{Latent Graph Inference using Projections of Model Spaces}
\label{sec: Latent Graph Inference using Projections of Model Spaces}
\vspace{-2mm}
With the three previously detailed model spaces, it is possible to get any value of the curvature in the range $(-\infty,\infty)$\footnote{This statement is not implying that there are positively curved hyperboloids or negatively curved hyperspheres, but rather that using the three aforementioned model spaces is enough to cover the full range of possible curvatures.}. However, a motivating factor to introduce projections of model spaces is the fact that points on the hyperboloid and hypersphere become divergent as $K_{\mathbb{S}}\rightarrow 0$ and $K_{\mathbb{H}}\rightarrow 0$. This effectively means that the manifolds become flat to stay consistent with their respective definitions. Furthermore, their distance and metric tensors do not become Euclidean at zero-curvature locations of the manifold, which may hinder curvature learning~\citep{Shopek2019}. We therefore consider the stereographic projections as suitable alternatives to these spaces, as they maintain a non-Euclidean structure and inherit many of the properties of the hyperboloid and the hypersphere.
\vspace{-2mm}
\subsection{The Poincaré Ball Model}
\vspace{-2mm}

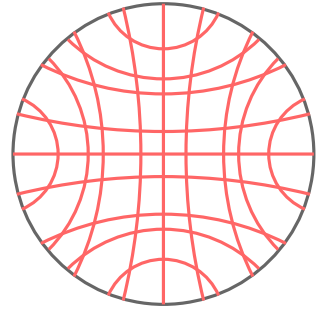
\begin{wrapfigure}{r}{5.5cm}
	\begin{tikzpicture}
		\filldraw[color=black!60, fill=white!5, very thick](0,0) circle (2);
		
		\draw[red!60, very thick] (0,-2) -- (0,2);
		\draw[red!60, very thick] (-2,0) -- (2,0);
		
		\filldraw[color=red!60, fill=none, very thick](-0.3,0)  arc (0:13.75:8.2) ;
		\filldraw[color=red!60, fill=none, very thick](-0.3,0)  arc (0:-13.75:8.2) ;
		\filldraw[color=red!60, fill=none, very thick](0.3,0)  arc (180:180+13.75:8.2) ;
		\filldraw[color=red!60, fill=none, very thick](0.3,0)  arc (180:180-13.75:8.2) ;
		
		\filldraw[color=red!60, fill=none, very thick](0,0.3)  arc (270:270+13.75:8.2) ;
		\filldraw[color=red!60, fill=none, very thick](0,0.3)  arc (270:270-13.75:8.2) ;
		\filldraw[color=red!60, fill=none, very thick](0,-0.3)  arc (90:90+13.75:8.2) ;
		\filldraw[color=red!60, fill=none, very thick](0,-0.3)  arc (90:90-13.75:8.2) ;

		\filldraw[color=red!60, fill=none, very thick](-1,0)  arc (0:44.9:1.8) ;
		\filldraw[color=red!60, fill=none, very thick](-1,0)  arc (0:-44.9:1.8) ;
		\filldraw[color=red!60, fill=none, very thick](1,0)  arc (180:180+44.9:1.8) ;
		\filldraw[color=red!60, fill=none, very thick](1,0)  arc (180:180-44.9:1.8) ;
		
		\filldraw[color=red!60, fill=none, very thick](0,1)
		arc (270:270+44.9:1.8) ;
		\filldraw[color=red!60, fill=none, very thick](0,1)  arc (270:270-44.9:1.8) ;
		\filldraw[color=red!60, fill=none, very thick](0,-1)  arc (90:90+44.9:1.8) ;
		\filldraw[color=red!60, fill=none, very thick](0,-1)  arc (90:90-44.9:1.8) ;
		
		\filldraw[color=red!60, fill=none, very thick](-0.8,0)  arc (0:26.9:3.6) ;
		\filldraw[color=red!60, fill=none, very thick](-0.8,0)  arc (0:-26.9:3.6) ;
		\filldraw[color=red!60, fill=none, very thick](0.8,0)  arc (180:180+26.9:3.6) ;
		\filldraw[color=red!60, fill=none, very thick](0.8,0)  arc (180:180-26.9:3.6) ;
		
		\filldraw[color=red!60, fill=none, very thick](0,0.8)  arc (270:270+26.9:3.6) ;
		\filldraw[color=red!60, fill=none, very thick](0,0.8)  arc (270:270-26.9:3.6) ;
		\filldraw[color=red!60, fill=none, very thick](0,-0.8)  arc (90:90+26.9:3.6) ;
		\filldraw[color=red!60, fill=none, very thick](0,-0.8)  arc (90:90-26.9:3.6) ;
		
		\filldraw[color=red!60, fill=none, very thick](-1.4,0)  arc (0:66.5:0.8) ;
		\filldraw[color=red!60, fill=none, very thick](-1.4,0)  arc (0:-66.5:0.8) ;
		\filldraw[color=red!60, fill=none, very thick](1.4,0)  arc (180:180+66.5:0.8) ;
		\filldraw[color=red!60, fill=none, very thick](1.4,0)  arc (180:180-66.5:0.8) ;
		
		\filldraw[color=red!60, fill=none, very thick](0, 1.4)  arc (270:270+66.5:0.8) ;
		\filldraw[color=red!60, fill=none, very thick](0,1.4)  arc (270:270-66.5:0.8) ;
		\filldraw[color=red!60, fill=none, very thick](0, -1.4)  arc (90:90+66.5:0.8) ;
		\filldraw[color=red!60, fill=none, very thick](0,-1.4)  arc (90:90-66.5:0.8) ;
		
	\end{tikzpicture}

	\caption{Grid of geodesics for the Poincaré ball.}
\end{wrapfigure} 
Hyperbolic space \citep{Cannon1997} is infinite; nevertheless, we can create a finite projection of hyperbolic space onto a flat surface. This is because, in principle, there are infinite many points within a disk. The Poincaré ball, also referred to as Poincaré disk in 2D, is a model of $n$-dimensional hyperbolic geometry which embeds all points in an $n$-dimensional hypersphere. It is a stereographic projection of the hyperboloid, meaning that it preserves angles but not distances. In particular, the distances between points grows exponentially as we move towards the outside of the Poincaré ball compared to their Euclidean distances. From here on, we will define the Poincaré ball as $\mathbb{P}_{K_\mathbb{P}}^{d_\mathbb{P}}$, where $K_\mathbb{P}$ denotes the curvature of the space and $d_\mathbb{P}$ its dimensionality. Distances between points in a Poincaré ball are given by  
\begin{equation}
	\mathfrak{d}_{\mathbb{P}}
	=
	\frac{1}{\sqrt{-K^\mathbb{P}}}
	\text{arccosh}
	\bigg(1-
	\frac{2K^\mathbb{P} ||\mathbf{x}-\mathbf{y}||_2^2}{(1+K^\mathbb{P} ||\mathbf{x}||_2^2)(1+K^\mathbb{P} ||\mathbf{y}||_2^2)}
	\bigg),
	\label{eq:poincare_distance}
\end{equation}
and its exponential map is in turn $exp_{\mathbf{x}_p}^{\mathbb{P}}(\mathbf{x})
=
\text{tanh}(\sqrt{-K^\mathbb{P}}||\mathbf{x}||_2)
\frac{\mathbf{x}}{\sqrt{-K^\mathbb{P}}||\mathbf{x}||_2}$
where the curvature $K_\mathbb{P}<0$. The negative curvature is inherited from the hyperboloid.  The GNN output at layer $l$ passes through a parametric transformation to obtain $\mathbf{\hat{x}}^{(l+1)}=f_{\mathbf{\Theta}}^{(l)}(\mathbf{x}^{(l)})$. $\mathbf{\hat{x}}^{(l+1)}$ is originally assumed to be on the tangent plane of the Poincaré manifold $\mathbb{P}_{K_\mathbb{P}}^{d_\mathbb{P}}$, so that $\mathbf{\hat{x}}^{(l+1)}\in\mathcal{T}_{p}\mathbb{P}_{K_\mathbb{P}}^{d_\mathbb{P}}$ and this is later projected to the actual manifold via the exponential map to obtain $\mathbf{\overline{x}}^{(l+1)} = exp_{\mathbf{x}_p}^{\mathbb{P}}(\mathbf{\hat{x}}^{(l+1)})\in\mathbb{P}_{K_\mathbb{P}}^{d_\mathbb{P}}$. The geodesic distance function in Equation~\ref{eq:poincare_distance} can finally be applied to the projected points.  Hence, the exponential map and distance function give us a principled way of mapping the latent features to the Poincaré ball.

\vspace{-3mm}
\subsection{The Stereographic Projection of the Hypersphere Model}
\vspace{-2mm}
The previous discussion also applies to the stereographic projection of the hypersphere. However, in this case the projected space has positive curvature. If we think about the lower-dimensional case in which we project a sphere onto a plane, in the polar aspect, the meridians project as straight lines originating at the pole and all angles between them remain truthful to those on the original manifold. On the other hand, parallels on the sphere would be shown as concentric circular arcs on the plane, with their spacing rapidly increasing from the pole, and distortion values remain the same along circular arcs surrounding the center point. The distance function for any arbitrary dimensionality is given by
\begin{equation}
	\mathfrak{d}_{\mathbb{D}}
	=
	\frac{1}{\sqrt{K^{\mathbb{D}}}}
	\text{arccos}
	\bigg(1-
	\frac{2K^{\mathbb{D}} ||\mathbf{x}-\mathbf{y}||_2^2}{(1+K^{\mathbb{D}} ||\mathbf{x}||_2^2)(1+K^{\mathbb{D}} ||\mathbf{y}||_2^2)}
	\bigg).
\end{equation}
The exponential map from a tangent Euclidean plane to the stereographic projection of the hypersphere is
$exp_{\mathbf{x}_p}^{\mathbb{D}}(\mathbf{x})
=
\text{tan}(\sqrt{K^{\mathbb{D}}}||\mathbf{x}||_2)
\frac{\mathbf{x}}{\sqrt{K^{\mathbb{D}}}||\mathbf{x}||_2}.$
\vspace{-2mm}
\section{Results}
\vspace{-2mm}

\label{sec: Results}
\vspace{-2mm}

In Table~\ref{table:homo_hetero_benchmarks}\footnote{The $*$ symbol after the dDGM$^{*}$ name denotes that the model does not use any graph provided by the dataset as an inductive bias. In other words, the dDGM$^{*}$ model generates the latent graph based solely on an input point cloud and its features $\mathbf{X}$. On the other hand, when the asterick is not present in the model name, the model uses the adjacency matrix $\mathbf{A}$ provided in the original dataset as inductive bias to generate the latent graph as well. The suffixes denote the use of different latent spaces: E~(Euclidean plane), H~(Hyperboloid), S~(Hypersphere), P~(Poincaré Ball), D~(Hypersphere Stereographic Projection). Double suffixes refer to product manifolds.} we display the results using both the Poincaré ball model and the stereographic projection of the hypersphere in isolation as embedding spaces, as well as part of product manifolds. We compare the results using the newly generated product manifolds against those previously obtained using the Euclidean plane, the hyperboloid, and the hypersphere for several homophilic, heterophilic, as well as real-world graph datasets. 

\begin{table}[h!]
	\caption{Results for benchmark homophilic, heterophilic, and real-world graph datasets. The \textbf{\textcolor{red}{First}}, \textbf{\textcolor{blue}{Second}}, and \textbf{\textcolor{brown}{Third}} best models for each dataset are highlighted in each table. $k$ denotes the number of connections per node when implementing the Gumbel Top-k sampling algorithm.}
	\centering 
	\scalebox{0.54}{
		\begin{tabular}{lcclcclc}
			\hline
			
			& \multicolumn{2}{c}{\textbf{HOMOPHILIC DATASETS}} & &\multicolumn{2}{c}{\textbf{HETEROPHILIC DATASETS}}& &\multicolumn{1}{c}{\textbf{REAL-WORLD DATASETS}}\\\hline
			
			& \textbf{Cora} & \textbf{CiteSeer} & &\textbf{Squirrel} & \textbf{Chameleon} & & \textbf{TadPole}\\ 
			Nodes & 2,708
			& 3,327 &  & 5,201 & 2,277 & & 564
			
			\\
			
			Features & 1,433
			& 3,703 &  & 2,089 & 2,325 & & 30
			\\

			Classes & 7
			& 6 & &  5 & 5 & & 3\\
			
			\hline
			\textbf{New Models} & \multicolumn{2}{c}{Accuracy  $(\%)$ $\pm$ Std} & \textbf{New Models} & \multicolumn{2}{c}{Accuracy  $(\%)$ $\pm$ Std}& \textbf{New Models} & \multicolumn{1}{c}{Accuracy  $(\%)$ $\pm$ Std}\\ 
			Connections $k$ & 
			
			7 &
			7 &
			Connections $k$ &3
			& 5   & Connections $k$ & 3
			
			\\ \hline
			
			\hline
			GCN-dDGM-P  &   $\textcolor{brown}{\boldsymbol{ 85.07 \pm 5.42 }}$ & $\boldsymbol{\textcolor{blue}{73.94 \pm 1.97}}$ & GCN-dDGM$^{*}$-P  & $ 21.38 \pm 2.66 $& $22.9 \pm 3.97$
			& GAT-dDGM$^{*}$-P & $90.54 \pm 4.23$ 
			\\
			GCN-dDGM-D &  $76.00 \pm 6.90$ & $71.98 \pm 4.09$& GCN-dDGM$^{*}$-D & $ 32.67 \pm 1.73 $& $46.91 \pm 3.11$ & GAT-dDGM$^{*}$-D & $57.32 \pm 11.33$
			
			\\
			
			\hline
			Connections $k$ & 
			
			5 &
			3 &
			Connections $k$ &3
			& 5  & Connections $k$ &3
			
			\\ \hline
			
			GCN-dDGM-HP  &  $\boldsymbol{\textcolor{brown}{85.07 \pm 3.41}}$ & $73.31 \pm 1.92$ & GCN-dDGM$^{*}$-HP  & $ 20.63 \pm 1.74$& $ 23.74 \pm 3.16$ & GAT-dDGM$^{*}$-HP & $86.25 \pm 6.29$
			
			\\
			GCN-dDGM-SD &  $83.96 \pm 3.52$ & $37.19 \pm 24.88$ &  GCN-dDGM$^{*}$-SD & $32.59 \pm 2.62$& $\boldsymbol{\textcolor{brown}{ 48.46 \pm 1.47}}$
			& GAT-dDGM$^{*}$-SD & $87.85 \pm 4.05$
			\\
			GCN-dDGM-EP  &  $85.07 \pm 4.46$ & $73.28 \pm 1.95$ & GCN-dDGM$^{*}$-EP  & $20.53 \pm 2.27$& $ 21.27 \pm 4.56$
			& GAT-dDGM$^{*}$-EP  & $88.21 \pm 4.80$
			\\
			GCN-dDGM-ED  &   $59.22 \pm 25.61$ &  $46.86 \pm 26.24$ & GCN-dDGM$^{*}$-ED  & $33.73 \pm 2.26$& $ 43.88 \pm 2.94$
			& GAT-dDGM$^{*}$-ED  &  $88.92 \pm 4.64$
			\\
			GCN-dDGM-HD  & $64.70 \pm 26.32$ &$50.81 \pm 27.48$ & GCN-dDGM$^{*}$-HD  & $21.03 \pm 1.95$& $ 21.94 \pm 4.04$
			& GAT-dDGM$^{*}$-HD  & $\boldsymbol{\textcolor{blue}{91.07 \pm 4.37}}$
			\\
			GCN-dDGM-EHP  &  $82.67 \pm 4.19$ &\textcolor{brown}{\boldsymbol{$73.82 \pm 2.64$}} & GCN-dDGM$^{*}$-EHP  & $21.73 \pm 3.06$& $22.16 \pm 3.48$
			& GAT-dDGM$^{*}$-EHP & $88.75 \pm 3.75$
			\\
			GCN-dDGM-EHD  &  $52.40 \pm 27.38$ & $30.30 \pm 21.17$ & GCN-dDGM$^{*}$-EHD  & $20.36 \pm 1.55$& $21.45 \pm 3.65$
			& GAT-dDGM$^{*}$-EHD  &  $90.71 \pm 3.36$
			\\
			GCN-dDGM-ESP  & $83.48 \pm 8.93$ & $73.55 \pm 1.70$ & GCN-dDGM$^{*}$-ESP  & $20.53 \pm 2.40$& $22.95 \pm 6.08$  &  GAT-dDGM$^{*}$-ESP & $90.00 \pm 4.67$
			\\
			GCN-dDGM-ESD  & $51.92 \pm 27.41$ & $25.21 \pm 16.82$ & GCN-dDGM$^{*}$-ESD  & $33.19 \pm 1.84$& $ 45.63 \pm 2.64$ & GAT-dDGM$^{*}$-ESD  & $89.82 \pm 6.34$
			\\
			GCN-dDGM-EPP  &  $\boldsymbol{\textcolor{blue}{85.96 \pm 4.01}}$& $73.79 \pm 3.11$ & GCN-dDGM$^{*}$-EPP  & $ 22.46 \pm 4.47$& $19.42 \pm 1.81$ & GAT-dDGM$^{*}$-EPP & $88.39 \pm 4.31$
			
			\\
			GCN-dDGM-EPD  &  $47.03 \pm 26.07$ & $47.56 \pm 26.51$ & GCN-dDGM$^{*}$-EPD  & $19.84 \pm 1.40$& $21.54 \pm 4.87$
			& GAT-dDGM$^{*}$-EPD  & $89.82 \pm 3.09$
			\\
			GCN-dDGM-EDD  &  $41.11 \pm 21.95$ & $31.11 \pm 22.91$ & GCN-dDGM$^{*}$-EDD  & $33.15 \pm 3.08$& $44.27 \pm 3.68$ & GAT-dDGM$^{*}$-EDD&  $\boldsymbol{\textcolor{blue}{91.07 \pm 3.09}}$
			\\
			GCN-dDGM-EHSPD  &   $30.03 \pm 3.16$ & $30.24 \pm 22.61$ & GCN-dDGM$^{*}$-EHSPD  & $19.48 \pm 1.17$& $21.18 \pm 4.29$ & GAT-dDGM$^{*}$-EHSPD  & $\boldsymbol{\textcolor{blue}{91.07 \pm 3.82}}$
			
			\\ \hline
			\textbf{Baselines} & \multicolumn{2}{c}{Accuracy $(\%)$ $\pm$ Std} & \textbf{Baselines} & \multicolumn{2}{c}{Accuracy $(\%)$ $\pm$ Std} & \textbf{Baselines} & \multicolumn{1}{c}{Accuracy $(\%)$ $\pm$ Std}  \\ 
			
			Connections $k$ & 
			
			7 &
			7 &
			Connections $k$ &3
			& 5   &  Connections $k$ &3
			
			\\ \hline
			
			GCN-dDGM$^{*}$-E  &   $82.11 \pm 4.24$  & $ 72.35 \pm 1.92$& GCN-dDGM$^{*}$-E  & $ 34.35 \pm 2.34$& $\textcolor{red}{\boldsymbol{48.90 \pm 3.61}}$ &  GAT-dDGM$^{*}$-E &$90.36 \pm 3.21$
			
			\\
			GCN-dDGM$^{*}$-H  &    $84.68  \pm 3.31$ & $70.43 \pm 4.95$& GCN-dDGM$^{*}$-H  & $ \textcolor{red}{\boldsymbol{35.00 \pm 2.35}}$ & $ 48.28 \pm 4.11$
			& GAT-dDGM$^{*}$-H &  $88.75 \pm 3.91$
			\\
			GCN-dDGM$^{*}$-S  &    $80.44 \pm 5.26$  &$72.89 \pm 2.00$& GCN-dDGM$^{*}$-S  & $ 33.12 \pm 2.22$& $ \textcolor{blue}{\boldsymbol{48.63 \pm 3.12}}$ &  GAT-dDGM$^{*}$-S &  $\textcolor{brown}{\boldsymbol{90.89 \pm 4.55}}$
			
			\\
			
			\hline
			Connections $k$ & 
			
			5 &
			3 &
			Connections $k$ &3
			& 5   &  Connections $k$ &3
			
			\\ \hline

			GCN-dDGM-HH  &  $76.09 \pm 7.11$ & $71.27 \pm 2.09$& GCN-dDGM$^{*}$-HH  &  $\textcolor{brown}{\boldsymbol{34.38 \pm 1.07}}$ & $48.33 \pm 4.14$
			& GAT-dDGM$^{*}$-HH & $89.68 \pm 5.70$
			\\
			GCN-dDGM-SS  &   $65.96 \pm 9.46$ & $59.16 \pm 5.96$& GCN-dDGM$^{*}$-SS  & $34.06 \pm 2.20$ & $48.28 \pm 3.07$ & GAT-dDGM$^{*}$-SS & $89.82 \pm 4.79$ \\
			GCN-dDGM-EH  &  $82.32 \pm 4.71$ & $72.89 \pm 1.64$& GCN-dDGM$^{*}$-EH  & $34.37 \pm 1.72$ & $47.58 \pm 3.85$ & GAT-dDGM$^{*}$-EH & $90.36 \pm 4.16$\\
			GCN-dDGM-ES  &  $81.44 \pm 5.80$& $71.87 \pm 3.20$& GCN-dDGM$^{*}$-ES  &  $33.38 \pm 1.86$ & $47.49 \pm 3.60$
			& GAT-dDGM$^{*}$-ES & $89.46 \pm 5.56$
			\\
			GCN-dDGM-HS  &   $82.59 \pm 4.50$ & $72.77 \pm 2.76$& GCN-dDGM$^{*}$-HS  & $\textcolor{blue}{\boldsymbol{34.65 \pm 2.45}}$ & $47.84 \pm 2.67$
			& GAT-dDGM$^{*}$-HS & $86.43 \pm 5.82$
			\\
			GCN-dDGM-EHH  &  $\textcolor{red}{\boldsymbol{86.63 \pm 3.25}}$ & \textcolor{red}{\boldsymbol{$75.42 \pm 2.39$}}& GCN-dDGM$^{*}$-EHH  & $33.19 \pm 1.92$ & $44.27 \pm 2.96$
			& GAT-dDGM$^{*}$-EHH & $87.68 \pm 9.95$
			\\
			GCN-dDGM-EHS  &$83.58 \pm 4.39$ & $69.98 \pm 2.70$ & GCN-dDGM$^{*}$-EHS  & $34.17 \pm 2.23$ & $47.58 \pm 4.54$ & GAT-dDGM$^{*}$-EHS & \textcolor{red}{\boldsymbol{$92.68 \pm 3.52$}}\\\hline
			GCN  & $83.11 \pm 2.29$& $69.97 \pm 2.06$& GCN  & $24.19 \pm 2.56$& $32.56 \pm 3.53$ 
			& GAT & N/A\\
			MLP  &  $58.92 \pm 3.28$ &  $59.48 \pm 2.14$& MLP  & $30.44 \pm 2.55$& $40.35 \pm 3.37$ & MLP & $87.68 \pm 3.52$ \\\hline
			
	\end{tabular}}
	
	\label{table:homo_hetero_benchmarks}
\end{table}

The stereographically projected models do sometimes outperform the original model spaces they were projected from and in general obtain comparable results. This is particularly the case for the homophilic datasets, Cora and Citeseer, and the TadPole dataset. However, in the case of Squirrel and Chameleon, for example, we do not observe any improvement with respect to the previous best-performing manifold signature combinations. Hence, we can conclude that using stereographic projections it is sometimes possible to outperform the original model space they were derived from while guaranteeing that the spaces do not become divergent as curvature tends to zero, and that their distance and metric tensors become Euclidean at zero-curvature locations of the manifold. More importantly, the introduction of these projections opens the possibility of constructing more diverse manifold signatures, while retaining the computational tractability of product manifolds.

\vspace{-4mm}
\section{Conclusion}
\vspace{-2mm}

\label{sec: Conclusion}

In this work we have proposed exploring the application of stereographically projected non-Euclidean model spaces for latent graph inference. These model spaces encode non-Euclidean geometries in a lower-dimensional manifold, and avoid divergence as their curvature tends to zero unlike their original counterparts. Effectively, these new projected spaces allow us to generate even more diverse product manifolds to try to infer the latent space of the data. However, given the large number of possible product manifold combinations, it has become apparent that future research should involve devising a principled way of inferring the signature of the product manifold. Furthermore, we would like to highlight that we use the Gumbel Top-k trick to be consistent with previous work by~\cite{Kazi_2022}. However, in general the Gumbel Top-k trick can be unstable to train and devising better edge sampling algorithms is left for future work.


\section{Acknowledgements}

AA thanks the Rafael del Pino Foundation for financial support. AA and HSOB thank the Oxford-Man Institute of Quantitative Finance for computing support.

\bibliography{iclr2023_conference}
\bibliographystyle{iclr2023_conference}

\end{document}